\documentclass[runningheads]{llncs}
\usepackage{graphicx}
\usepackage{algorithm}
\usepackage{algpseudocode}
\usepackage{mathptmx}
\usepackage{multirow}
\usepackage[normalem]{ulem}
\usepackage{vcell}
\usepackage{booktabs}
\makeatletter
\newcommand{\printfnsymbol}[1]{%
  \textsuperscript{\@fnsymbol{#1}}%
}

\makeatother
\begin{document}
\title{A Heuristic-driven Ensemble Framework for COVID-19 Fake News Detection}
%
%
\author{Sourya Dipta Das\inst{1}\thanks{equal contribution} \and Ayan Basak\inst{1}\printfnsymbol{1} \and Saikat Dutta\inst{2}}
\authorrunning{Sourya et al.}
%
\institute{Razorthink Inc, USA \\
\email{\{souryadipta.das,ayan.basak\}@razorthink.com} \and
IIT Madras, India\\
\email{cs18s016@smail.iitm.ac.in}}
\maketitle              
\begin{abstract}
The significance of social media has increased manifold in the past few decades as it helps people from even the most remote corners of the world stay connected. With the COVID-19 pandemic raging, social media has become more relevant and widely used than ever before, and along with this, there has been a resurgence in the circulation of fake news and tweets that demand immediate attention. In this paper, we describe our Fake News Detection system that automatically identifies whether a tweet related to COVID-19 is “real” or “fake”, as a part of CONSTRAINT COVID19 Fake News Detection in English challenge. We have used an ensemble model consisting of pre-trained models that has helped us achieve a joint 8\textsuperscript{th} position on the leader board. We have achieved an F1-score of 0.9831 against a top score of 0.9869. Post completion of the competition, we have been able to drastically improve our system by incorporating a novel heuristic algorithm based on username handles and link domains in tweets fetching an F1-score of 0.9883 and achieving state-of-the art results on the given dataset.
\keywords{COVID-19  \and Language Model \and Fake News \and Ensemble \and Heuristic.}
\end{abstract}
\section{Introduction}
Fake news represents the press that is used to spread false information and hoaxes through conventional platforms as well as online ones, mainly social media. There has been an increasing interest in fake news on social media due to the political climate prevailing in the modern world \cite{Tucker et al.,Calvillo et al.,Monti et al.}, as well as several other factors. Detecting misinformation on social media is as important as it is technically challenging. The difficulty is partly due to the fact that even humans cannot accurately distinguish false from true news, mainly because it involves tedious evidence collection as well as careful fact checking. With the advent of technology and ever-increasing propagation of fake articles in social media, it has become really important to come up with automated frameworks for fake news identification. In this paper, we describe our system which performs a binary classification on tweets from social media and classifies it into “real” or “fake”. We have used transfer learning in our approach as it has proven to be extremely effective in text classification tasks, with a reduced training time as we do not need to train each model from scratch. The primary steps for our approach initially include text preprocessing, tokenization, model prediction, and ensemble creation using a soft voting schema. Post evaluation, we have drastically improved our fake news detection framework with a heuristic post-processing technique that takes into account the effect of important aspects of tweets like username handles and URL domains. This approach has allowed us to produce much superior results when compared to the top entry in the official leaderboard. We have performed an ablation study of the various attributes used in our post-processing approach. We have also provided examples of tweets where the post-processing approach has predicted correctly when compared to the initial classification output.

\section{Related Work}
Traditional machine learning approaches have been quite successful in fake news identification problems. Reis et al. \cite{Reis et al.} has used feature engineering to generate hand-crafted features like syntactic features, semantic features etc. 
The problem was then approached as a binary classification problem where these features were fed into conventional Machine Learning classifiers like K-Nearest Neighbor (KNN), Random Forest (RF), Naive Bayes, Support Vector Machine (SVM) and XGBOOST (XGB), where RF and XGB yielded results that were quite favourable.
Shu et al. \cite{Shu et al.} have proposed a novel framework TriFN, which provides a principled way to model tri-relationship among publishers, news pieces, and users simultaneously. This framework significantly outperformed the baseline Machine Learning models as well as erstwhile state-of-the-art frameworks. With the advent of deep learning, there has been a significant revolution in the field of text classification, and thereby in fake news detection. Karimi et al. \cite{Karimi et al.} has proposed a Multi-Source Multi-class Fake News Detection framework that can do automatic feature extraction using Convolution Neural Network (CNN) based models and combine these features coming from multiple sources using an attention mechanism, which has produced much better results than previous approaches that involved hand-crafted features. Zhang et al. \cite{Zhang et al.} introduced a new diffusive unit model, namely Gated Diffusive Unit (GDU), that has been used to build a deep diffusive network model to learn the representations of news articles, creators and subjects simultaneously. 
Ruchansky et al. \cite{Ruchansky et al.} has proposed a novel CSI(Capture-Score-Integrate) framework that uses an Long Short-term Memory (LSTM) network to capture the temporal spacing of user activity and a doc2vec \cite{doc2vec} representation of a tweet, along with a neural network based user scoring module to classify the tweet as real or fake. It emphasizes the value of incorporating all three powerful characteristics in the detection of fake news: the tweet content, user source, and article response.
Monti et al. \cite{Monti et al.} has shown that social network structure and propagation are important features for fake news detection by implementing  a geometric deep learning framework using Graph Convolutional Networks.	

\textbf{Language models:} Most of the current state-of-the-art language models are based on Transformer\cite{Vaswani et al.} and they have proven to be highly effective in text classification problems. They provide superior results when compared to previous state-of-the-art approaches using techniques like Bi-directional LSTM, Gated Recurrent Unit (GRU) based models etc. The models are trained on a huge corpus of data. The introduction of the BERT \cite{Devlin et al.} architecture has transformed the capability of transfer learning in Natural Language Processing. It has been able to achieve state-of-the art results on downstream tasks like text classification. RoBERTa \cite{Liu et al.} is an improved version of the BERT model. It is derived from BERT’s language-masking strategy, modifying its key hyperparameters, including removing BERT’s next-sentence pre-training objective, and training with much larger mini-batches and learning rates, leading to improved performance on downstream tasks. XLNet \cite{Yang et al.} is a generalized auto-regressive language method. It calculates the joint probability of a sequence of tokens based on the transformer architecture having recurrence. Its training objective is to calculate the probability of a word token conditioned on all permutations of word tokens in a sentence, hence capturing a bidirectional context. XLM-RoBERTa \cite{Conneau et al.} is a transformer \cite{Vaswani et al.} based language model relying on Masked Language Model Objective. DeBERTa \cite{He et al.} provides an improvement over the BERT and RoBERTa models using two novel techniques; first, the disentangled attention mechanism, where each word is represented using two vectors that encode its content and position, respectively, and the attention weights among words are computed using disentangled matrices on their contents and relative positions, and second, the output softmax layer is replaced by an enhanced mask decoder to predict the masked tokens pre-training the model. ELECTRA \cite{Clark et al.} is used for self-supervised language representation learning. It can be used to pre-train transformer networks using very low compute, and is trained to distinguish "real" input tokens vs "fake" input tokens, such as tokens produced by artificial neural networks. ERNIE 2.0 \cite{Sun et al.} is a continual pre-training framework to continuously gain improvement on knowledge integration through multi-task learning, enabling it to learn various lexical, syntactic and semantic information through massive data much better.

\section{Dataset Description}
The dataset \cite{Patwa et al.} for CONSTRAINT COVID-19 Fake News Detection in English challenge was provided by the organizers on the competition website\footnote{https://competitions.codalab.org/competitions/26655}. It consists of data that have been collected from various social media and fact checking websites, and the veracity of each post has been verified manually. The “real” news items were collected from verified sources which give useful information about COVID-19, while the “fake” ones were collected from tweets, posts and articles which make speculations about COVID-19 that are verified to be false. The original dataset contains 10,700 social media news items, the vocabulary size (i.e., unique words) of which is 37,505 with 5141 words in common to both fake and real news. It is class-wise balanced with 52.34\% of the samples consisting of real news, and 47.66\% of fake samples. These are 880 unique username handle and 210 unique URL domains in the data.

\section{Methodology}
We have approached this task as a text classification problem. Each news item needs to be classified into two distinct categories: “real” or “fake”. Our proposed method consists of five main parts: (a) Text Preprocessing, (b) Tokenization, (c) Backbone Model Architectures, (d) Ensemble, and (e) Heuristic Post Processing. The overall architecture of our system is shown in Figure-\ref{initial_process_diag_fig}. More detailed description is given in the following subsections.

\subsection{Text Preprocessing}
Some social media items, like tweets, are mostly written in colloquial language. Also, they contain various other information like usernames, URLs, emojis, etc. We have filtered out such attributes from the given data as a basic preprocessing step, before feeding it into the ensemble model. We have used the tweet-preprocessor\footnote{pypi.org/project/tweet-preprocessor/} library from Python to filter out such noisy information from tweets.

\subsection{Tokenization}
During tokenization, each sentence is broken down into tokens before being fed into a model. We have used a variety of tokenization approaches\footnote{huggingface.co/docs/tokenizers/python/latest/} depending upon the pre-trained model that we have used, as each model expects tokens to be structured in a particular manner, including the presence of model-specific special tokens. Each model also has its corresponding vocabulary associated with its tokenizer, trained on a large corpus data like GLUE, wikitext-103, CommonCrawl data etc. During training, each model applies the tokenization technique with its corresponding vocabulary on our tweets data.
We have used a combination of XLNet \cite{Yang et al.}, RoBERTa \cite{Liu et al.}, XLM-RoBERTa \cite{Conneau et al.}, DeBERTa \cite{He et al.}, ERNIE 2.0 \cite{Sun et al.} and ELECTRA \cite{Clark et al.} models and have accordingly used the corresponding tokenizers from the base version of their pre-trained models.

\subsection{Backbone Model Architectures}
We have used a variety of pre-trained language models\footnote{huggingface.co/models} as backbone models for text classification. For each model, an additional fully connected layer is added to its respective encoder sub-network to obtain prediction probabilities for each class- ``real" and ``fake" as a prediction vector. We have used transfer learning in our approach in this problem. Each model has used some pre-trained model weights as initial weights. Thereafter, it fine-tunes the model weights using the tokenized training data. The same tokenizer is used to tokenize the test data and the fine-tuned model checkpoint is used to obtain predictions during inference. 

\subsection{Ensemble}
In this method, we use the model prediction vectors from the different models to obtain our final classification result, i.e. “real” or “fake”. To balance an individual model's limitations, an ensemble method can be useful for a collection of similarly well-performing models. We have experimented with two approaches: soft voting and hard voting, that are described in the following figure:

\begin{figure}
\centering
\includegraphics[width=\textwidth]{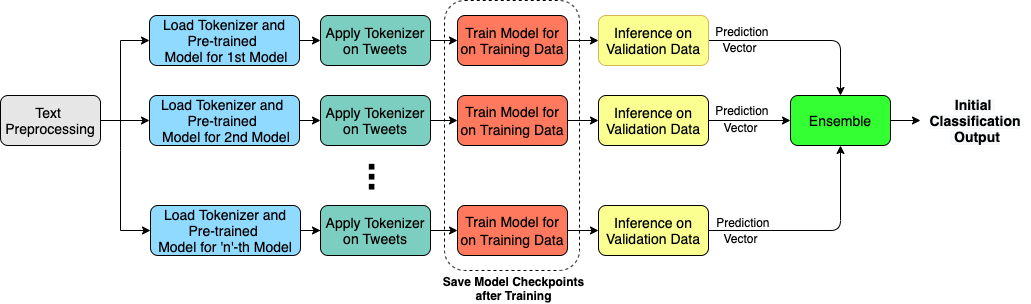}
\caption{Fake News Identification Initial Process Block Diagram} 
\label{initial_process_diag_fig}
\end{figure}

\subsubsection{Soft Voting : }
In this approach, we calculate a “soft probability score” for each class by averaging out the prediction probabilities of various models for that class. The class that has a higher average probability value is selected as the final prediction class. 
Probability for ``real" class, $P^r(x)$ and probability for ``fake" class , $P^f(x)$ for a tweet $x$ is given by, 
\begin{equation}\label{equ1}
    P^r(x) = \sum_{i=1}^{n}\frac{P^r_i(x)}{n}
\end{equation}
\begin{equation}\label{equ2}
   P^f(x) = \sum_{i=1}^{n}\frac{P^f_i(x)}{n}
\end{equation}
where $P^r_i(x)$ and $P^f_i(x)$ are ``real" and ``fake" probabilities by the $i$-th model and $n$ is the total number of models. 
\subsubsection{Hard Voting : } 
In this approach, the predicted class label for a news item is the class label that represents the majority of the class labels predicted by each individual model. In other words, the class with the most number of votes is selected as the final prediction class. Votes for ``real" class, $V^r(x)$ and Votes for ``fake" class , $V^f(x)$ for a tweet $x$ is given by,
\begin{equation}\label{equ3}
   V^r(x) = \sum_{i=1}^{n}{I(P^r_i(x) \geq P^f_i(x))}
\end{equation}
\begin{equation}\label{equ4}
   V^f(x) = \sum_{i=1}^{n}{I(P^r_i(x) < P^f_i(x))}
\end{equation}
where the value of $I(a)$ is $1$ if condition $a$ is satisfied and $0$ otherwise. 
\subsection{Heuristic Post-Processing}
In this approach, we have augmented our original framework with a heuristic approach that can take into account the effect of username handles and URL domains present in some data, like tweets. This approach works well for data having URL domains and username handles; we rely only on ensemble model predictions for texts lacking these attributes. We create a new feature-set using these attributes. Our basic intuition is that username handles and URL domains are very important aspects of a tweet and they can convey reliable information regarding the genuineness of tweets. We have tried to incorporate the effect of these attributes along with our original ensemble model predictions by calculating probability vectors corresponding to both of them. We have used information about the frequency of each class for each of these attributes in the training set to compute these vectors. In our experiments, we observed that Soft-voting works better than Hard-voting. Hence our post-processing step takes Soft-voting prediction vectors into account. The steps taken in this approach are described as follows:

\begin{itemize}
    \item   First, we obtain the class-wise probability from the best performing ensemble model. These probability values form two features of our new feature-set.
    
    \item We collect username handles from all the news items in our training data, and calculate how many times the ground truth is “real” or “fake” for each username.
    
    \item   We calculate the conditional probability of a particular username indicating a real news item, which is represented as follows:
        
        \begin{equation}\label{equ3}
            P^r(x|username) = \frac{n(A)}{n(A) + n(B)}
        \end{equation}
        
    where n(A) = number of “real” news items containing the username and  n(B) = number of “fake” news items containing the username. Similarly, the conditional probability of a particular username indicating a fake news item is given by,
        
        \begin{equation}\label{equ4}
            P^f(x|username) = \frac{n(B)}{n(A) + n(B)}
        \end{equation}

        We obtain two probability vectors that form four additional features of our new dataset.

\begin{figure}[b]
\centering
\includegraphics[width=\textwidth]{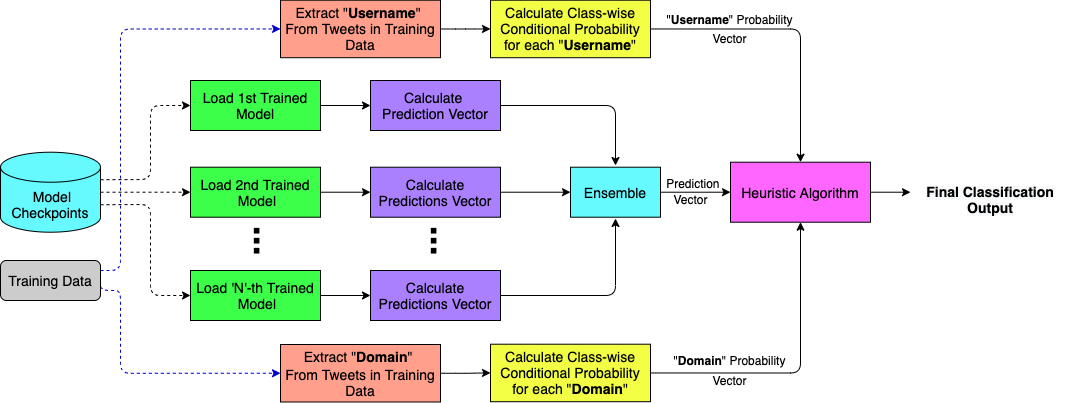}
\caption{Fake News Identification Post Process Block Diagram} 
\label{post_process_diag_fig}
\end{figure}

\item  We collect URL domains from all the news items in our training data, obtained by expanding the shorthand URLs associated with the tweets, and calculate how many times the ground truth is “real” or “fake” for each domain.

\item  We calculate the conditional probability of a particular URL domain indicating a real news item, which is represented as follows:
        
        \begin{equation}\label{equ3}
            P^r(x|domain) = \frac{n(P)}{n(P) + n(Q)}
        \end{equation}
        
    where n(P) = number of “real” news items containing the domain and  n(Q) = number of “fake” news items containing the domain. Similarly, the conditional probability of a particular domain indicating a fake news item is given by,
        \begin{equation}\label{equ4}
            P^f(x|domain) = \frac{n(Q)}{n(P) + n(Q)}
        \end{equation}
            We obtain two probability vectors that form the final two additional features of our new dataset.
            
\item  In case there are multiple username handles and URL domains in a sentence, the final probability vectors are obtained by averaging out the vectors of the individual attributes.

\item   At this point, we have new training, validation and test feature-sets obtained using class-wise probability vectors from ensemble model outputs as well as probability values obtained using username handles and URLs from the training data. We use a novel heuristic algorithm on this resulting feature set to obtain our final class predictions.

\end{itemize}
Table \ref{username domain examples} shows some samples of the conditional probability values of each label class given each of the two attributes, URL domain and username handle. We have also shown the frequency of those attributes in the training data.
\begin{table}
\centering
\caption{Few Examples on URL Domain-name and Username attribute distribution data }
\label{username domain examples}
\resizebox{\textwidth}{!}{%
\begin{tabular}{|c|c|c|c|l|c|c|c|c|} 
\cline{1-4}\cline{6-9}
\multicolumn{4}{|c|}{Example of URL Domain Name Prob. Dist.} & ~ ~ & \multicolumn{4}{c|}{Example of UserName Prob. Dist.} \\ 
\cline{1-4}\cline{6-9}
\begin{tabular}[c]{@{}c@{}}URL Domain \\Name \end{tabular} & \begin{tabular}[c]{@{}c@{}}$P^r(x|domain)$ \end{tabular} & \begin{tabular}[c]{@{}c@{}}$P^f(x|domain)$ \end{tabular} & \begin{tabular}[c]{@{}c@{}}Frequency \end{tabular} & \multirow{6}{*}{} & UserName & \begin{tabular}[c]{@{}c@{}}$P^r(x|username)$ \end{tabular} & \begin{tabular}[c]{@{}c@{}}$P^f(x|username)$ \end{tabular} & \begin{tabular}[c]{@{}c@{}}Frequency \end{tabular} \\ 
\cline{1-4}\cline{6-9}
news.sky & 1.0 & 0.0 & 274 &  & MoHFW\_NDIA & 0.963 & 0.037 & 162 \\ 
\cline{1-4}\cline{6-9}
medscape.com & 1.0 & 0.0 & 258 &  & DrTedros & 1.0 & 0.0 & 110 \\ 
\cline{1-4}\cline{6-9}
thespoof.com & 0.0 & 1.0 & 253 &  & ICMRDELHI & 0.9903 & 0.0097 & 103 \\ 
\cline{1-4}\cline{6-9}
newsthump.com & 0.0 & 1.0 & 68 &  & PIB\_ndia & 1.0 & 0.0 & 83 \\ 
\cline{1-4}\cline{6-9}
theguardian.com & 0.167 & 0.833 & 6 &  & CDCMMWR & 1.0 & 0.0 & 34 \\
\cline{1-4}\cline{6-9}
\end{tabular}
}
\end{table}
The details of the heuristic algorithm is explained in the following pseudocode (Algorithm-\ref{h_algo}). In our experiment, the value of threshold used is 0.88. The post-processing architecture is shown in Figure-\ref{post_process_diag_fig}.
\begin{algorithm}
	\caption{Heuristic Algorithm}
	\label{h_algo}
	{\textbf{Result: }label ( ``real" or ``fake") }
\begin{algorithmic}[1]
  \If {$P^r(x|username) > threshold $ AND $P^r(x|username) > P^f(x|username)$} 
    \State {label = ``real"}
  \ElsIf{$P^f(x|username) > threshold $ AND $P^r(x|username) < P^f(x|username)$  }
    \State {label = ``fake"}
  \ElsIf {$P^r(x|domain) > threshold $ AND $P^r(x|domain) > P^f(x|domain)$} 
    \State {label = ``real"}
  \ElsIf{$P^f(x|domain) > threshold $ AND $P^r(x|domain) < P^f(x|domain)$  }
    \State {label = ``fake"}
  \ElsIf{$P^r(x) > P^f(x)$}
    \State {label = ``real"}
    \Else 
    \State {label = ``fake"}
  \EndIf
\end{algorithmic}
\end{algorithm}
\section{Experiments and Results}
\subsection{System Description}
We have fine-tuned our pre-trained models using AdamW\cite{adamW} optimizer and cross-entropy loss after doing label encoding on the target values. We have applied softmax on the logits produced by each model in order to obtain the prediction probability vectors. The experiments were performed on a system with 16GB RAM and 2.2 GHz Quad-Core Intel Core i7 Processor, along with a Tesla T4 GPU, with batch size of 32. The maximum input sequence length was fixed at 128. Initial learning rate was set to 2e-5. The number of epochs varied from 6 to 15 depending on the model.

\subsection{Performance of Individual Models}
We have used each fine-tuned model individually to perform “real” vs “fake” classification. Quantitative results are tabulated in Table-\ref{single_base_model}. We can see that XLM-RoBERTa, RoBERTa, XLNet and ERNIE 2.0 perform really well on the validation set. However, RoBERTa has been able to produce the best classification results when evaluated on the test set.

\begin{table}[h]
\centering
\caption{Individual model performance on validation and test set}
\label{single_base_model}
\resizebox{\textwidth}{!}{%
\begin{tabular}{|c|c|c|c|c|c|c|c|c|}
\hline
\multicolumn{1}{|c|}{\multirow{2}{*}{\textbf{Model Name}}} & \multicolumn{4}{c|}{\textbf{Validation Set}} & \multicolumn{4}{c|}{\textbf{Test set}} \\ \cline{2-9} 
\multicolumn{1}{|c|}{} & \multicolumn{1}{c|}{\textbf{Accuracy}} & \multicolumn{1}{c|}{\textbf{Precision}} & \multicolumn{1}{c|}{\textbf{Recall}} & \multicolumn{1}{c|}{\textbf{F1 Score}} & \multicolumn{1}{c|}{\textbf{Accuracy}} & \multicolumn{1}{c|}{\textbf{Precision}} & \multicolumn{1}{c|}{\textbf{Recall}} & \multicolumn{1}{c|}{\textbf{F1 Score}} \\ \hline
XLM-RoBERTa (base) & 0.968 & 0.968 & 0.968 & 0.968 & 0.970 & 0.970 & 0.970 & 0.970 \\ \hline
RoBERTa (base) & 0.970 & 0.970 & 0.970 & 0.970 & \textbf{0.972} & \textbf{0.972} & \textbf{0.972} & \textbf{0.972} \\ \hline
XLNet (base, cased) & 0.975 & 0.975 & 0.975 & 0.975 & 0.966 & 0.966 & 0.966 & 0.966 \\ \hline
DeBERTa (base) & 0.964 & 0.964 & 0.964 & 0.964 & 0.964 & 0.964 & 0.964 & 0.964 \\ \hline
ELECTRA (base) & 0.948 & 0.948 & 0.948 & 0.948 & 0.953 & 0.953 & 0.953 & 0.953 \\ \hline
ERNIE 2.0 & \textbf{0.976} & \textbf{0.976} & \textbf{0.976} & \textbf{0.976} & 0.969 & 0.969 & 0.969 & 0.969 \\ \hline
\end{tabular}%
}
\end{table}
\vspace{-10px}
\subsection{Performance of Ensemble Models}
We tried out different combinations of pre-trained models with both the ensemble techniques: Soft Voting and Hard Voting. Performance for different ensembles are shown in Table-\ref{soft_voting_model} and \ref{hard_voting_model}.
From the results, we can infer that the ensemble models significantly outperform the individual models, and Soft-voting ensemble method performed better overall than Hard-voting ensemble method. Hard-voting Ensemble model consisting of RoBERTa, XLM-RoBERTa, XLNet, ERNIE 2.0 and DeBERTa models performed the best among other hard voting ensembles on both validation and test set. Among the Soft Voting Ensembles, the ensemble consisting of RoBERTa, XLM-RoBERTa, XLNet, ERNIE 2.0 and Electra models achieved best accuracy overall on the validation set and 
a combination of XLNet, RoBERTa, XLM-RoBERTa and DeBERTa models produces the best classification result overall on the test set. Our system has been able to achieve an overall F1-score of 0.9831 and secure a joint 8\textsuperscript{th} rank in the leaderboard, against a top score of 0.9869.

\begin{table}[h]
\centering
\caption{Performance of Soft Voting for different ensemble models on validation and test set}
\label{soft_voting_model}
\resizebox{\textwidth}{!}{%
\begin{tabular}{|c|c|c|c|c|c|c|c|c|} 
\hline
\multirow{2}{*}{\begin{tabular}[c]{@{}c@{}}\textbf{ Ensemble Model}\\\textbf{ Combination} \end{tabular}} & \multicolumn{4}{c|}{\textbf{Validation Set} } & \multicolumn{4}{c|}{\textbf{Test set} } \\ 
\cline{2-9}
 & \textbf{Accuracy}  & \textbf{Precision}  & \textbf{Recall}  & \textbf{F1 Score}  & \textbf{Accuracy}  & \textbf{Precision}  & \textbf{Recall}  & \textbf{F1 Score}  \\ 
\hline
\begin{tabular}[c]{@{}c@{}} RoBERTa+XLM-RoBERTa\\ +XLNet\\ \end{tabular} & 0.9827 & 0.9827 & 0.9827 & 0.9827 & 0.9808 & 0.9808 & 0.9808 & 0.9808 \\ 
\hline
\begin{tabular}[c]{@{}c@{}} RoBERTa+XLM-RoBERTa\\ +XLNet+DeBERT\\ \end{tabular} & 0.9832 & 0.9832 & 0.9832 & 0.9832 & \textbf{0.9831} & \textbf{0.9831} & \textbf{0.9831} & \textbf{0.9831} \\ 
\hline
\begin{tabular}[c]{@{}c@{}} RoBERTa+XLM-RoBERTa\\ +XLNet+ERNIE 2.0\\ +DeBERTa\\ \end{tabular} & 0.9836 & 0.9836 & 0.9836 & 0.9836 & 0.9822 & 0.9822 & 0.9822 & 0.9822 \\ 
\hline
\begin{tabular}[c]{@{}c@{}} RoBERTa+XLM-RoBERTa\\ +XLNet+ERNIE 2.0\\ +Electra\\ \end{tabular} & \textbf{0.9841} & \textbf{0.9841} & \textbf{0.9841} & \textbf{0.9841} & 0.9808 & 0.9808 & 0.9808 & 0.9808 \\
\hline
\end{tabular}
}
\end{table}
\vspace{-20px}
\begin{table}[h]
\centering
\caption{Performance of Hard Voting for different ensemble models on validation and test set}
\label{hard_voting_model}
\resizebox{\textwidth}{!}{%
\begin{tabular}{|c|c|c|c|c|c|c|c|c|} 
\hline
\multirow{2}{*}{\begin{tabular}[c]{@{}c@{}}\textbf{ Ensemble Model}\\\textbf{ Combination} \end{tabular}} & \multicolumn{4}{c|}{\textbf{Validation Set} } & \multicolumn{4}{c|}{\textbf{Test set} } \\ 
\cline{2-9}
 & \textbf{Accuracy}  & \textbf{Precision}  & \textbf{Recall}  & \textbf{F1 Score}  & \textbf{Accuracy}  & \textbf{Precision}  & \textbf{Recall}  & \textbf{F1 Score}  \\ 
\hline
\begin{tabular}[c]{@{}c@{}} RoBERTa+XLM-RoBERTa\\ +XLNet\\ \end{tabular} & 0.9818 & 0.9818 & 0.9818 & 0.9818 & 0.9804 & 0.9804 & 0.9804 & 0.9804 \\ 
\hline
\begin{tabular}[c]{@{}c@{}} RoBERTa+XLM-RoBERTa\\ +XLNet+DeBERT\\ \end{tabular} & 0.9748 & 0.9748 & 0.9748 & 0.9748 & 0.9743 & 0.9743 & 0.9743 & 0.9743 \\ 
\hline
\begin{tabular}[c]{@{}c@{}} RoBERTa+XLM-RoBERTa\\ +XLNet+ERNIE 2.0\\ +DeBERTa\\ \end{tabular} & \textbf{0.9832} & \textbf{0.9832} & \textbf{0.9832} & \textbf{0.9832} & \textbf{0.9813} & \textbf{0.9813} & \textbf{0.9813} & \textbf{0.9813} \\ 
\hline
\begin{tabular}[c]{@{}c@{}} RoBERTa+XLM-RoBERTa\\ +XLNet+ERNIE 2.0\\ +Electra\\ \end{tabular} & 0.9822 & 0.9822 & 0.9822 & 0.9822 & 0.9766 & 0.9766 & 0.9766 & 0.9766 \\
\hline
\end{tabular}
}
\end{table}
\vspace{-17px}
\subsection{Performance of Our Final Approach}
We augmented our Fake News Detection System with an additional heuristic algorithm and achieved an overall F1-score of 0.9883, making this approach state-of-the-art on the given fake news dataset \cite{Patwa et al.}. We have used the best performing ensemble model consisting of RoBERTa, XLM-RoBERTa, XLNet and DeBERTa for this approach. We have shown the comparison of the results on the test set obtained by our model before and after applying the post-processing technique against the top 3 teams in the leaderboard in Table \ref{post proc Comp}. Table \ref{result_tab} shows a few examples where the post-processing algorithm corrects the initial prediction. The first example is corrected due to extracted domain which is \textit{``news.sky"} and the second one is corrected because of presence of the username handle, \textit{``@drsanjaygupta"}. 
\begin{table}
\centering
\caption{Performance comparison on test set}
\label{post proc Comp}
\begin{tabular}{|c|c|c|c|c|} 
\hline
 \textbf{Method}  & \textbf{Accuracy}  & \textbf{Precision}  & \textbf{Recall}  & \textbf{F1 Score}  \\ 
\hline
Team g2tmn (\textit{Rank 1}) & 0.9869 & 0.9869 & 0.9869 & 0.9869 \\ 
\hline
Team saradhix (\textit{Rank 2}) & 0.9864 & 0.9865 & 0.9864 & 0.9864 \\ 
\hline
Team xiangyangli (\textit{Rank 3}) & 0.9860 & 0.9860 & 0.9860 & 0.9860 \\ 
\hline
\begin{tabular}[c]{@{}c@{}}Ensemble Model\\ \end{tabular} & 0.9831 & 0.9831 & 0.9831 & 0.9831 \\ 
\hline
\begin{tabular}[c]{@{}c@{}}Ensemble Model +\\Heuristic Post-Processing \end{tabular} & \textbf{0.9883} & \textbf{0.9883} & \textbf{0.9883} & \textbf{0.9883} \\
\hline
\end{tabular}
\end{table}
\vspace{-17px}
\begin{table}
\caption{Qualitative comparison between our initial and final approach.}
\centering
\label{result_tab}
\resizebox{\textwidth}{!}{%
\footnotesize
\begin{tabular}{lccc}
\multicolumn{1}{c}{\uline{Tweet}} & \begin{tabular}[c]{@{}c@{}}\uline{Initial }\\\uline{Classification }\\\uline{Output}\end{tabular} & \begin{tabular}[c]{@{}c@{}}\uline{Final }\\\uline{Classification }\\\uline{Output}\end{tabular} & \uline{Ground Truth} \\ 
\hline
\begin{tabular}[c]{@{}l@{}}Coronavirus: Donald Trump ignores COVID-19 rules \\with 'reckless and selfish' indoor rally https://t.co/JsiHGLMwfO\end{tabular} & fake & real & real \\ 
\hline
\begin{tabular}[c]{@{}l@{}}We're LIVE talking about COVID-19 (a vaccine transmission)\\ with @drsanjaygupta. Join us and ask some questions\\ of your own: https://t.co/e16G2RGdkA https://t.co/Js7lemT1Z6\end{tabular} & real & fake & fake \\ 
\hline
\end{tabular}
}
\label{example_pred}
\end{table}
\vspace{-22px}
\subsection{Ablation Study}
We have performed an ablation study by assigning various levels of priority to each of the features (username and domain) and then checking which class’s probability value for that feature is maximum for a particular tweet, so that we can assign the corresponding “real” or “fake” class label to that particular tweet. For example, in one iteration, we have given URL domains a higher priority than username handles to select the label class. We have also experimented with only one attribute mentioned above in our study. Results for different priority and feature set is shown in Table \ref{ablation_tab}.

Another important parameter that we have introduced for our experiment is a threshold on the class-wise probability values for the features. For example, if the probability that a particular username that exists in a tweet belongs to “real” class  is greater than that of it belonging to “fake” class, and the probability of it belonging to the “real” class is greater than a specific threshold, we assign a “real” label to the tweet. The value of this threshold is a hyperparameter that has been tuned based on the classification accuracy on the validation set. We have summarized the results from our study with and without the threshold parameter in Table \ref{ablation_tab}.\newline
As we can observe from the results, domain plays a significant role for ensuring a better classification result when the threshold parameter is taken into account. The best results are obtained when we consider the threshold parameter and both the username and domain attributes, with a higher importance given to the username. 

\begin{table}[h]
\centering
\caption{Ablation Study on Heuristic algorithm}
\label{ablation_tab}
\resizebox{\textwidth}{!}{%
\begin{tabular}{|c|c|c|c|c|} 
\hline
\multirow{2}{*}{\begin{tabular}[c]{@{}c@{}}\textbf{Combination of Attributes~}\\(\textbf{in descending order of Attribute Priority}) \end{tabular}} & \multicolumn{2}{c|}{\textbf{with Threshold} } & \multicolumn{2}{c|}{\textbf{without Threshold} } \\ 
\cline{2-5}
 & \begin{tabular}[c]{@{}c@{}}\textbf{F1 Score on}\\\textbf{Validation Set}\end{tabular} & \begin{tabular}[c]{@{}c@{}}\textbf{F1 Score on}\\\textbf{Test Set}\\ \end{tabular} & \begin{tabular}[c]{@{}c@{}}\textbf{F1 Score on}\\\textbf{Validation Set}\\ \end{tabular} & \begin{tabular}[c]{@{}c@{}}\textbf{F1 Score on}\\\textbf{Test Set}\\ \end{tabular} \\ 
\hline
\{\textit{username, ensemble model pred}~\}  & 0.9831 & 0.9836 & \textbf{0.9822} & \textbf{0.9804} \\ 
\hline
\{\textit{domain, ensemble model pred~}\}  & \textbf{0.9917} & 0.9878 & 0.9635 & 0.9523 \\ 
\hline
\{\textit{domain, username, ensemble model pred~}\}  & 0.9911 & 0.9878 & 0.9635 & 0.9519 \\ 
\hline
\{\textit{username, domain, ensemble model pred~}\}  & 0.9906 & \textbf{0.9883} & 0.9645 & 0.9528 \\
\hline
\end{tabular}
}
\label{Post_Proc_scores}
\end{table}

\vspace{-19px}

\section{Conclusion}
In this paper, we have proposed a robust framework for identification of fake tweets related to COVID-19, which can go a long way in eliminating the spread of misinformation on such a sensitive topic. In our initial approach, we have tried out various pre-trained language models. Our results have significantly improved when we implemented an ensemble mechanism with Soft-voting by using the prediction vectors from various combinations of these models. Furthermore, we have been able to augment our system with a novel heuristics-based post-processing algorithm that has drastically improved the fake tweet detection accuracy, making it state-of-the-art on the given dataset. Our novel heuristic approach shows that username handles and URL domains form very important features of tweets and analyzing them accurately can go a long way in creating a robust framework for fake news detection. 
Finally, we would like to pursue more research into how other pre-trained models and their combinations perform on the given dataset. It would be really interesting to evaluate how our system performs on other generic Fake News datasets and also if different values of the threshold parameter for our post-processing system would impact its overall performance.

%
%
%

\end{document}